\pdfoutput=1
\documentclass{article}
\usepackage{iclr2027_conference,times}

\usepackage{amsmath,amsfonts,bm}

\def\eqref#1{equation~\ref{#1}}

\def\1{\bm{1}}

\DeclareMathAlphabet{\mathsfit}{\encodingdefault}{\sfdefault}{m}{sl}
\SetMathAlphabet{\mathsfit}{bold}{\encodingdefault}{\sfdefault}{bx}{n}

\usepackage{hyperref}
\usepackage{url}
\usepackage{booktabs}
\usepackage{graphicx}
\usepackage{amsmath}
\usepackage{float}
\usepackage{adjustbox}

\newcommand{\OTSbaseCircuitOne}{0.530}
\newcommand{\OTSbaseCircuitPeak}{0.642}
\newcommand{\OTSbaseCircuitFar}{0.583}
\newcommand{\OFTbaseCircuitPeak}{0.619}

\newcommand{\OTSsurgCircuitTwo}{0.500}
\newcommand{\OTSsurgCircuitThree}{0.058}

\newcommand{\OFTsurgCircuitTwo}{0.090}

\newcommand{\OTSbaseHopsPeak}{0.166}

\newcommand{\OFTbaseHopsPeak}{0.154}
\newcommand{\OTSsurgHopsEight}{0.000}
\newcommand{\OTSsurgProbePaired}{0.556}
\newcommand{\OFTsurgProbePaired}{0.485}
\newcommand{\OTSbaseProbePaired}{0.674}
\newcommand{\EntBaseCircuitOne}{0.786}
\newcommand{\EntSurgCircuitTwo}{12.470}
\newcommand{\EntSurgCircuitSixteen}{12.267}
\newcommand{\EntSurgCircuitFar}{5.672}
\newcommand{\MarginSurgBudget}{0.263}
\newcommand{\MarginSurgFar}{1.247}
\newcommand{\NormRatioMax}{2.021}

\newcommand{\OTSsurgProbeFloor}{0.519}

\newcommand{\OFTsurgProbeFloor}{0.525}

\newcommand{\OTSbaseProbeFloor}{0.555}

\newcommand{\OTSbaseHopsProbeMajority}{0.072}
\newcommand{\ScrOffKOne}{0.790}
\newcommand{\ScrOffKTwo}{0.948}
\newcommand{\ScrOffKFour}{0.999}
\newcommand{\ScrOffKEight}{0.997}
\newcommand{\ScrOffKSixteen}{0.981}
\newcommand{\ScrOffKThirtyTwo}{0.932}
\newcommand{\ScrOnKOne}{0.492}
\newcommand{\ScrOnKTwo}{0.713}
\newcommand{\ScrOnKFour}{0.963}
\newcommand{\ScrOnKEight}{0.772}
\newcommand{\ScrOnKSixteen}{0.618}
\newcommand{\ScrOnKThirtyTwo}{0.584}

\newcommand{\ScrSeedsA}{16}
\newcommand{\ScrSeedsB}{6}
\newcommand{\ScrDeltaMax}{0.867}
\newcommand{\ScrPKOne}{2.2\times10^{-7}}

\newcommand{\ScrPKFour}{0.012}

\newcommand{\ScrBeyondCells}{3}
\newcommand{\ScrBeyondHolm}{3}
\newcommand{\ScrBeyondPMax}{1.1\times10^{-5}}
\newcommand{\ScrBeyondPMin}{4.0\times10^{-6}}
\newcommand{\BdeepFar}{0.907}
\newcommand{\BmidFar}{0.838}
\newcommand{\BwideFar}{0.490}
\newcommand{\ScrTwoOffKOne}{0.979}

\newcommand{\ScrTwoOffKFour}{0.999}
\newcommand{\ScrTwoOffKEight}{0.999}

\newcommand{\ScrTwoOnKOne}{0.556}

\newcommand{\ScrTwoOnKFour}{0.988}
\newcommand{\ScrTwoOnKEight}{0.946}

\newcommand{\ScrTwoSeeds}{6}
\newcommand{\ScrTwoSeedsA}{16}
\newcommand{\ScrTwoDeltaMax}{0.844}
\newcommand{\ScrTwoPMin}{8.9\times10^{-6}}
\newcommand{\ScrTwoBeyondCells}{3}
\newcommand{\ScrTwoBeyondHolm}{2}

\newcommand{\ScrTwoPKOne}{2.3\times10^{-6}}

\newcommand{\ScrTwoPKFour}{0.361}

\newcommand{\ScrTwoPKSixteen}{9.0\times10^{-4}}
\newcommand{\ScrTwoPKThirtyTwo}{0.080}
\newcommand{\BTwoMidDelta}{0.528}
\newcommand{\BTwoMidP}{0.180}
\newcommand{\BTwoWideDelta}{0.778}
\newcommand{\BTwoWideP}{0.026}

\newcommand{\BTwoDeepMultEight}{0.957}

\newcommand{\BTwoMidMultEight}{0.733}

\newcommand{\BTwoWideMultEight}{0.737}

\newcommand{\BOneDeepMultEight}{0.907}

\newcommand{\BOneMidMultEight}{0.918}

\newcommand{\BOneWideMultEight}{0.828}
\newcommand{\RefitOneBase}{0.998}
\newcommand{\RefitOneSurg}{0.639}
\newcommand{\RefitOneRefit}{0.756}
\newcommand{\RefitOneSeeds}{6}
\newcommand{\RefitOneBaseSixteen}{0.621}

\newcommand{\RefitOneFrac}{33}
\newcommand{\RefitOnePSurg}{0.031}
\newcommand{\RefitOnePBase}{0.031}

\newcommand{\RefitTwoSurg}{0.775}
\newcommand{\RefitTwoRefit}{0.845}

\newcommand{\RefitTwoFrac}{31}
\newcommand{\RefitTwoPSurg}{0.062}
\newcommand{\RefitTwoPBase}{0.062}
\newcommand{\TransferMax}{0.030}
\newcommand{\TransferMin}{0.001}
\newcommand{\TransferFloor}{0.125}
\newcommand{\TransferCells}{3}
\newcommand{\FloorCircuit}{0.500}
\newcommand{\FloorHops}{0.038}
\newcommand{\DoseBaseBudget}{0.642}

\newcommand{\DoseKTwoBudget}{0.496}

\newcommand{\DoseEntBaseBudget}{0.95}

\newcommand{\DoseEntKTwoBudget}{3.51}

\newcommand{\DoseEntKFourBudget}{12.11}

\newcommand{\DosePKTwoBudget}{3.1\times10^{-9}}

\newcommand{\DoseCells}{3}

\newcommand{\RefitCurveHeadNormThreeHundred}{33}

\newcommand{\RefitCurveHeadNormThreeThousand}{35}

\newcommand{\RefitCurveHeadNormLastThirty}{43}

\newcommand{\RefitCurveHeadNormLastThreeThousand}{100}
\newcommand{\RefitOneSurgOne}{0.578}
\newcommand{\RefitOneRefitOne}{0.671}

\newcommand{\PairedTier}{tier A}
\newcommand{\PairedOTScircuitGapBase}{-0.101}
\newcommand{\PairedOTScircuitGapSurg}{+0.490}
\newcommand{\PairedOTScircuitDelta}{-0.591}
\newcommand{\PairedOTScircuitCI}{[-0.609, -0.567]}
\newcommand{\PairedOTScircuitP}{1.0\times10^{-140}}
\newcommand{\PairedOTScircuitN}{800}

\newcommand{\PairedOFTcircuitGapBase}{-0.011}
\newcommand{\PairedOFTcircuitGapSurg}{+0.516}

\newcommand{\PairedOTShopsDelta}{-0.110}

\newcommand{\PairedOTShopsN}{900}

\title{Agreement rather than normalisation keeps depth reusable in looped language
models}

\author{Ivan Viakhirev$^{1}$\quad Kirill Borodin$^{2}$\quad Maxim Abramov$^{3}$\quad Grach Mkrtchian$^{1}$\\[4pt]
\normalfont $^{1}$lab260, Yerevan, Armenia\\
\normalfont $^{2}$BitmanagerAI, Dubai, UAE\\
\normalfont $^{3}$St.~Petersburg Federal Research Center of the Russian Academy of Sciences,\\
\normalfont \phantom{$^{3}$}St.~Petersburg, Russia\\[4pt]
\normalfont Correspondence to: \texttt{kborodin.research@gmail.com}}

\iclrfinalcopy

\begin{document}

\maketitle
\lhead{Preprint. Under review at ICLR 2027}

\begin{abstract}
Looped language models apply one block of layers repeatedly, and the design pays off only
if depth is reusable: loop steps beyond the trained budget must remain usable at inference.
We ask whether that rests on normalising the state between loop steps, as the released
Ouro models do, or on agreement between the loop a model is trained with and the loop it
is run with. On released weights this loop-boundary normalisation can be removed without
adding parameters and with the readout path bit-identical, and the operation is an identity
at one loop step, which serves as an implementation check. Removing it takes Ouro-1.4B and
Ouro-2.6B to zero accuracy by the trained budget on a balanced binary task (chance
\FloorCircuit{}). State-norm growth, the failure that prior accounts predict, is not the
cause: the norm stays within a factor of \NormRatioMax{} while the head commits to a single
wrong token. Yet in 19M-parameter models trained from scratch on two synthetic tasks,
\ScrSeedsA{} seeds per configuration, training without the normalisation is never worse
than training with it and extrapolates better: the two are closest at the trained depth and
separate on both sides of it (Cliff's $\delta$ up to $\ScrDeltaMax$ beyond the budget).
Both results follow if agreement, and not the normalisation itself, is what keeps depth
reusable. The damage sits in the readout together with the last layer of the block: in the
models trained from scratch, refitting the head over the frozen modified loop recovers a
third of the lost accuracy and plateaus, while refitting the last block layer with it
recovers all of it. Finally, comparing looped with stacked depth at an equal inference loop
count is confounded by unequal overshoot of the trained budget; at matched overshoot the
advantage of looping vanishes on one task and shrinks on the other.
\end{abstract}

\section{Introduction}

Looped language models apply one stack of layers to the state $r$ times, with $r$ chosen
at inference and expected to remain useful beyond the number of loop steps $r_0$ used in
training. Without that, looping only saves parameters at a fixed depth. Which part of the
architecture keeps loop steps reusable past the trained budget, and which part ties the
model to $r_0$, is usually left open.

Two released models happen to differ in just this detail. Both normalise
inside their recurrent block, but Ouro \citep{zhu2025ouro} also renormalises the state at
the end of every loop step, with the module that normalises the readout input, while
Huginn \citep{geiping2025} applies that module once per forward pass. We call this
placement the loop-boundary normalisation and determine it by instrumenting the models
(Sec.~\ref{sec:axis}); it makes an architectural hypothesis testable without training.
Restricting Ouro's model-level normalisation to the final loop step leaves the readout
path bit-identical and changes only the state that re-enters the loop, adds no
parameters, and is an identity at $r=1$, which checks the implementation.

One rearrangement cannot separate two explanations. The first is that boundary
normalisation is required, because without it the state diverges and the loop stops being
reusable. The second is that agreement is required, because the readout is calibrated on
the states that training produces, and any change to the loop topology moves its input
off that distribution, in either direction. On released weights both predict the same
outcome, since the training topology there is fixed, so we repeat the experiment on
models trained from scratch with the topology set before training: fifty models per task,
\ScrSeedsA{} seeds per configuration on the normalisation axis, with the training seed as
the unit of analysis and rank tests over seeds.

The answer has two parts, and the two halves of the study supply one each. Removing
boundary normalisation from a model trained with it breaks it within a few loop steps, so
agreement is necessary. Training from scratch without the normalisation is never worse
than training with it and is better at every depth other than the trained one, so given
agreement the normalisation is a cost and not a benefit: the configuration trained with
it peaks at its trained depth and degrades on both sides, although its training and
inference loops agree. Neither part is what the literature predicts. The nearest prior
account expects an architecture with a normalisation between loop steps to avoid the
failure by construction, and the residual-growth account expects failure through the
scale of the state. The state norm of the damaged model changes by at most a factor of
\NormRatioMax{} over the whole grid, while readout entropy rises from \EntBaseCircuitOne{}
to \EntSurgCircuitTwo{} bits on a one-bit task and then falls back as the head commits to
a single token outside the answer alphabet.

Our contributions are the following.

\begin{enumerate}
\item An exact rearrangement of the loop-boundary normalisation on released weights
  (\mbox{Huginn-3.5B}, \mbox{Ouro-1.4B}, \mbox{Ouro-2.6B}), a measured topology audit, an
  identity check at $r=1$, and measurements of readout entropy, logit margin, state norm
  per loop step and input-to-prediction mutual information (Sec.~\ref{sec:r1}).
\item The causal version of the same question on models trained from scratch with the
  topology fixed before training, on two task families, \ScrSeedsA{} seeds per
  configuration, an exact rank test and Cliff's $\delta$ (Sec.~\ref{sec:r3}).
\item A separation of ``normalisation is required'' from ``the training and inference
  topologies must agree'' in favour of the second, on four measurements: the state
  scale barely moves while accuracy collapses, the
  readout leaves the answer alphabet and then collapses onto a constant, a readout refit
  over a frozen loop plateaus at a third of the loss while refitting the last block layer
  with it returns all of it, and training without a boundary normalisation is better
  although both configurations agree with their own loop
  (Sec.~\ref{sec:r2}--\ref{sec:refit}).
\item A methodological result on comparing depth in the loop against depth in the stack:
  comparing at equal absolute $K$ favours whichever configuration has the larger trained
  loop count $r_0$, although it is the one with less inference compute; at equal budget
  overshoot $K/r_0$, which is also equal compute, the advantage of looping is not
  detectable on \texttt{circuit} and shrinks without closing on \texttt{hops}
  (Sec.~\ref{sec:r4}).
\end{enumerate}

\section{Related work}

\paragraph{Looped models and depth as an inference-time resource.}
Reusing one block instead of a stack goes back to the Universal Transformer
\citep{dehghani2019}, and \citet{giannou2023} showed that a looped transformer can
simulate arbitrary programs at a constant parameter count. Three recent systems raised this to
language-model scale: Huginn-3.5B \citep{geiping2025}, a prelude, core and coda topology
trained on 800B tokens; Ouro \citep{zhu2025ouro}, which moves the loop into pretraining on
7.7T tokens with an entropy regulariser on the depth distribution and a learned exit gate;
and retrofitted recurrence \citep{mcleish2025}, which converts an already pretrained model.
Around these sit elastic-depth variants \citep{jeddi2026loopformer}, training-free wrappers
\citep{chen2026trainingfree}, memory-decoupled implementations \citep{conchello2026memory},
scaling studies \citep{huang2026latentrecurrent,gao2026loopies} and analyses of looping as
a substitute for chain of thought \citep{kohli2026loopthink,fan2026bridging,zhang2026memorybudget}.

\paragraph{Where the normalisation sits.}
\citet{blayney2026mechanistic} tabulate the block equations of the released models:
Huginn normalises the residual stream after each sublayer while Ouro normalises the
attention and MLP outputs, and both apply a normalisation after the loop. Normalisation
inside a loop has since been analysed as an implicit gain control on a local Lipschitz
constant \citep{buehlmaier2026layernorm}, as the condition for reachable, input-dependent
fixed points \citep{labovich2026stability}, through pre-norm against post-norm placement
\citep{movahedi2026fixedpoint}, and through which iterates converge at all
\citep{zhang2026convergence,logan2026pertoken,feinashley2026attractor}. Past the trained
loop count, \citet{kuo2026stabilizing} trace out-of-distribution variance to a spurious
correlation between sequence length and loop count, and halting gates and trajectory
readouts are diagnosable in their own right \citep{popescu2026adaptive,lam2026energyentropy}.
None of this work holds the readout fixed and moves the loop-boundary normalisation on
released weights.

\paragraph{The nearest neighbour.}
The closest work is on the readout blind spot \citep{sharma2026readout}: dense
per-iteration cross entropy supervises only the state variables visible through a
scale-invariant readout, so a pre-norm residual recurrence keeps updating the radial scale
unseen by the loss, and at 44M and 129M parameters the final state norm reaches thousands.
Its intervention on released weights is a scalar rescaling of Ouro-1.4B's final hidden
state before the readout, not a change of loop topology. That work varies the readout
normalisation and holds the loop-boundary normalisation fixed; we do the reverse, so the
two studies cover orthogonal axes of one design space. Its limitations section names our
condition as open, expecting inter-loop normalisation to avoid the failure by
construction; we measure that case, and training with a boundary normalisation makes
extrapolation worse (Sec.~\ref{sec:r3}). Both that account and the residual-growth account
predict failure through state-norm growth, which our measurements do not support: the
state scale barely moves while the readout leaves the answer alphabet and then settles
on a constant.

\paragraph{Depth in the loop against depth in the stack.}
\citet{schwethelm2026isodepth} fit a joint scaling law over an iso-depth pretraining
sweep and report a recurrence-equivalence exponent of $0.46$, so shared recurrences raise
validation loss at matched training compute, and \citet{oncescu2026recurrent} approach the
same trade-off from effective depth and decoding cost. Both read off one quality number at
fixed training compute or effective depth; we ask the inference-time question, comparing
three placements of a matched depth against the loop count used at inference
(Sec.~\ref{sec:r4}).

\paragraph{Abrupt emergence of in-context lookup.}
Induction-head formation shows up as a bump in the training curve \citep{olsson2022},
and associative recall with many candidate keys formalises the harder variant
\citep{arora2023zoology}; we observe such a transition on single-hop lookup over an
in-context map (App.~\ref{app:scope}).

\section{Setup}

\subsection{Models}
\label{sec:models}

The released checkpoints, Huginn-3.5B, Ouro-1.4B and Ouro-2.6B, carry the scale; the
causal claims rest on models trained from scratch, from 19M parameters at six layers to
76M at twenty-four, where the topology is a flag set before training. The division is
deliberate: sixteen independent training runs per configuration are not feasible at 3B
parameters, and without them there is no rank test over seeds, so the two halves answer
different questions (Table~\ref{tab:models}). We write $r$ for the inference-time loop count of released models
and $K$ for that of models trained from scratch; $r_0$ is the trained budget in both cases.

\subsection{The axis: loop-boundary normalisation}
\label{sec:axis}

By \emph{loop-boundary normalisation} we mean a normalisation applied to the state that
leaves one loop iteration and enters the next. It is distinct from the normalisations
inside the block, which both released models have and which \citet{blayney2026mechanistic}
tabulate.

We determine the placement by instrumentation. Every normalisation module in a loaded
model is instrumented and its calls are counted over one forward pass at two loop counts;
a module whose count grows with the loop count acts inside the loop, and a module called
the same number of times acts on the readout path. On released weights the audit returns
\texttt{norm\_in\_loop} for Ouro, with the model-level normalisation that produces the
readout input called once per loop step, and \texttt{readout\_only} for Huginn. Since
the instrument counts calls to normalisation \emph{modules}, \texttt{readout\_only} means that no such module is called
per loop step, not that no normalisation acts there; this matters only for the Huginn
direction (Sec.~\ref{sec:limits}).

Two operations follow. For Ouro we restrict the model-level
normalisation to the final loop step, so the readout path is bit-identical and only the
state re-entering the loop changes; this is the operation our results rest on. For Huginn
we apply the readout-path normalisation to the state before it re-enters the block
(Sec.~\ref{sec:r1} and Sec.~\ref{sec:limits}). In the models trained from scratch the same
axis is a flag set before training, with a dedicated normalisation applied between
iterations and skipped on the final one; unlike the released case, that normalisation has
its own parameters instead of sharing them with the readout. At one loop step every one
of these operations is an identity by construction, and we use that identity check as a
gate: an arm whose $r=1$ numbers differ from its baseline is discarded.

\subsection{Tasks}

We use three families of synthetic tasks with an exact per-instance depth $d^\star$
and an independent width control $w$. The answer is a single token, so arg-max metrics
are defined without decoding. The family \texttt{circuit} evaluates a Boolean circuit
of depth $d^\star$; \texttt{hops} follows $d^\star$ pointers through an in-context map
with 26 possible answers; \texttt{paragg} computes the parity of a subset of answers.
Instances are generated on a grid of cells $(d^\star, w)$; a \emph{stratum} is one cell
and a \emph{split} the evaluation instances of a family. The grid is materialised in two
tiers, 100 instances per cell over the full grid and more on the pre-registered primary
cell. The generator and the verifier are separate code, an instance that fails
verification does not enter a split, and both binary families are balanced by
construction at 50/50 in every stratum.

\subsection{Metrics and statistics}
\label{sec:metrics}

Accuracy is the arg-max over the model's full vocabulary compared with the gold answer
token, reported against an explicit floor equal to the larger of random guessing within
the answer alphabet and the best constant answer: \FloorCircuit{} for the binary families
and, for \texttt{hops}, \FloorHops{} on released models, which answer over the 26-letter
alphabet, and $1/w$ on models trained from scratch on a single cell of width $w$; where
the majority answer of a split beats chance, that majority floor is reported instead. The
floor is defined within the alphabet and the arg-max is not, which matters for one
reading: on a balanced binary family a constant answer within the alphabet scores
\FloorCircuit{}, so an accuracy of zero means that the top token has left the answer
alphabet, and the format statistics of App.~\ref{app:collapse} keep that case separable
from a wrong answer within it. We also report readout entropy in
bits, the top-1 minus top-2 logit, the state norm per loop step, and, for collapse, the
mutual information between input and arg-max prediction with the Miller-Madow correction,
the modal share and the number of distinct predictions. On released weights we use an exact
McNemar test on paired instances with a cluster bootstrap over (family, $d^\star$, data
seed) and a Holm correction; on models trained from scratch the unit of analysis is the
training seed, \ScrSeedsA{} per configuration on the normalisation axis and \ScrSeedsB{}
on the depth axis, with an exact Mann-Whitney test and Cliff's $\delta$; averaging over
instances would treat one run as a thousand observations.

A competence gate is fixed in advance: a pair of arm and family enters the analysis
only if peak accuracy reaches 0.30 in absolute terms and exceeds the floor by more than
0.05, and the pairs that fail it are reported as such.

\section{Results}

\subsection{The rearrangement is asymmetric}
\label{sec:r1}

\begin{figure}[t]
\centering
\includegraphics[width=\textwidth]{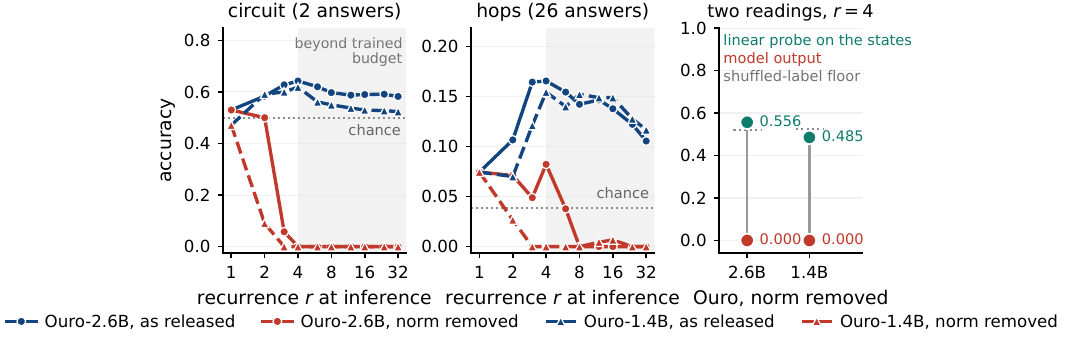}
\caption{Rearranging the loop-boundary normalisation on released weights, four
in-context examples in the prompt; blue as released, red with the boundary normalisation
removed, line style marks the model. Left and middle: accuracy against the inference-time
loop count $r$; the shaded band lies beyond the trained budget $r_0=4$ and the dotted
line is chance inside the answer alphabet (on \texttt{hops} the majority floor,
\OTSbaseHopsProbeMajority{}, is higher; the $y$ ranges differ). Right: the same states of
the arm with the normalisation removed, at the trained budget on \texttt{circuit}, read
by the model's own head and by a linear probe fitted to them, with the floor from a probe
fitted to shuffled labels. The probe sits at its floor while the output sits at zero,
below anything an answer within the alphabet would give (Sec.~\ref{sec:r2}).}
\label{fig:asymmetry}
\end{figure}

The identity check passes exactly: at $r=1$ the difference in accuracy and in readout
entropy between an arm and its rearranged counterpart is $0.0000$ in all nine pairs of arm
and family zero-shot and all four few-shot, so everything at $r>1$ is a property of the
loop and not of weight loading or answer scoring.

Removing the boundary normalisation from Ouro breaks the model within the first loop
steps. On \texttt{circuit} with four in-context examples, Ouro-2.6B goes from
\OTSbaseCircuitOne{} at $r=1$ to \OTSsurgCircuitTwo{} at $r=2$, \OTSsurgCircuitThree{} at
$r=3$ and zero, to three decimals, from the trained budget onward, while the unmodified
model peaks at \OTSbaseCircuitPeak{} at $r=4$ and still reads \OTSbaseCircuitFar{} at
$r=32$ (Fig.~\ref{fig:asymmetry}; full grid in Table~\ref{tab:erelacc}). Ouro-1.4B
breaks earlier, at \OFTsurgCircuitTwo{} at $r=2$ and zero at $r=3$, so the damage is
complete within two to three loop steps. On \texttt{hops} the same arm reaches
\OTSsurgHopsEight{} by $r=8$ while the unmodified model peaks at \OTSbaseHopsPeak{}
against the majority floor of \OTSbaseHopsProbeMajority{} (chance over the alphabet is
\FloorHops{}).

The operation destroys a plateau. The unmodified models peak at the trained budget and
decline slowly, from \OTSbaseCircuitPeak{} to \OTSbaseCircuitFar{} on Ouro-2.6B, and the
claim of this section is about what preserves that plateau. The reverse operation, adding
a boundary normalisation to Huginn, is reported in Table~\ref{tab:huginn} but carries no
claim, since Huginn fails the competence gate on all three families before anything is
done to it. The asymmetry we rely on is between removing the normalisation from released
weights and training without it from scratch (Sec.~\ref{sec:r3}).

The predictions collapse onto a single token along with the accuracy: at $r=32$ on \texttt{hops} the unmodified
arms spread their answers over 25 and 30 distinct tokens, while with the boundary
normalisation removed Ouro-1.4B returns a single token for the whole split at zero mutual
information with the input (App.~\ref{app:collapse}).

\subsection{The readout collapses; the state scale does not}
\label{sec:r2}

\begin{figure}[t]
\centering
\includegraphics[width=\textwidth]{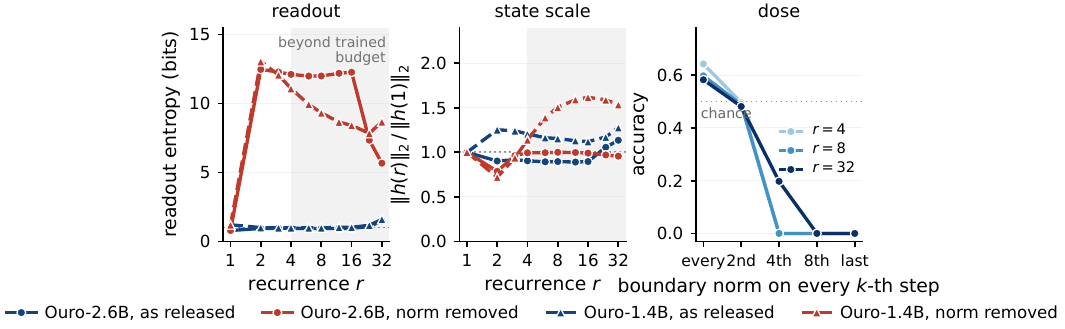}
\caption{Mechanism on \texttt{circuit}; colours and line styles as in
Fig.~\ref{fig:asymmetry}, shaded band beyond the trained budget. Left: readout entropy in
bits, dotted line at the one bit of a uniform guess over two answers. Middle: state norm
at $r$ over state norm at $r=1$. Right: Ouro-2.6B with the boundary normalisation applied
every $k$-th loop step and always on the last, at three depths; the response is a
threshold. The arms with the normalisation removed lose the readout while their state
scale stays within a factor of \NormRatioMax{}.}
\label{fig:mechanism}
\end{figure}

The literature on instability in looped models predicts a specific failure mode: with
no boundary normalisation the residual stream grows without bound and the model fails
because of the scale of the state. The prediction is not borne out on released weights.
Across every cell of the grid and both arms with the operation applied, the ratio of the
state norm at $r$ to its value at $r=1$ stays below \NormRatioMax{}, and the absolute norm
stays in the range of the unmodified arms (Fig.~\ref{fig:mechanism}, middle).

The readout does change, in two stages. Entropy first rises, from
\EntBaseCircuitOne{} bits at $r=1$ to \EntSurgCircuitTwo{} bits at $r=2$, and stays
near \EntSurgCircuitSixteen{} bits up to $r=16$, which on a vocabulary of roughly 15
bits is close to uniform over the whole vocabulary. In this stage the zero accuracy is
the arg-max of a nearly flat head landing outside a two-token alphabet
(Sec.~\ref{sec:metrics}): the head has stopped answering in format. Entropy then
comes back down, to \EntSurgCircuitFar{} bits at $r=32$, but not towards the correct
answer: over the same cells the top-1 minus top-2 logit grows from \MarginSurgBudget{} at
the trained budget to \MarginSurgFar{} at $r=32$ (Fig.~\ref{fig:margin} for both
families), the modal share reaches 0.83, and accuracy stays at zero. A modal share of 0.83
with zero accuracy on a balanced binary family is possible only if the modal token lies
outside the answer alphabet. The head first loses the ability to separate candidates and
then becomes confident in a constant that is not an answer. Neither stage is explained by the scale of
the state, which stays within a factor of \NormRatioMax{} of its value at $r=1$ over the
same range.

A linear probe shows what kind of failure this is. We fit a fresh linear probe to the
states of the same arm at the trained budget $r_0=4$ and read it on the same evaluation
instances, so the two readings differ only in the head. On \texttt{circuit} the output
of Ouro-2.6B with the boundary normalisation removed is zero while the probe on those
states reads at chance (\OTSsurgProbePaired{} against a shuffled-label floor of
\OTSsurgProbeFloor{}; on Ouro-1.4B, \OFTsurgProbePaired{} against \OFTsurgProbeFloor{}),
so the answer is not recoverable from the state by a linear map either. At the trained
budget, then, the state no longer carries the answer linearly and the head, nearly
uniform over the vocabulary, no longer answers within the alphabet; the commitment to a
constant comes later, by $r=32$. On the unmodified arm the same procedure gives
\OTSbaseProbePaired{} against a floor of \OTSbaseProbeFloor{}, slightly above that arm's
own output of \OTSbaseCircuitPeak{}: both readings find the answer, and the state carries
at least as much of it as the trained head extracts.

The operation also admits a dose. Normalising every second, fourth or eighth loop step,
always including the last so that the readout path is unchanged, orders the inference
topologies by distance from the trained one, and agreement predicts accuracy that is
non-increasing along that order. It is, at all \DoseCells{} depths (Table~\ref{tab:dose}):
at the trained budget the model as released reads \DoseBaseBudget{}, every second step
\DoseKTwoBudget{} (McNemar $p=\DosePKTwoBudget$) and every fourth or sparser zero, with
readout entropy of \DoseEntBaseBudget{}, \DoseEntKTwoBudget{} and \DoseEntKFourBudget{}
bits on a one-bit task. The response is a threshold, not a gradient: halving the frequency
of the normalisation leaves the head at chance, anything sparser takes it to zero, out of
the answer alphabet, and the state norm stays in range throughout.

The pre-registered paired test used the gap between the probe and the model output and
predicted a small gap for Ouro as released and a large one after the boundary
normalisation is removed. The gap moves from small to large, reversing sign, in all four
pairs of arm and family, with $\Delta=\PairedOTScircuitDelta$ on \texttt{circuit} for
Ouro-2.6B (McNemar $p=\PairedOTScircuitP$, $n=\PairedOTScircuitN$), and all four survive
the Holm correction (App.~\ref{app:paired}). Since the model output after the operation
is zero, the gap there equals the probe accuracy: the test measures the distance between
a probe at chance and a head that has left the answer alphabet.

\subsection{Trained from scratch: the normalisation hurts away from the trained depth}
\label{sec:r3}

\begin{figure}[t]
\centering
\includegraphics[width=0.9\textwidth]{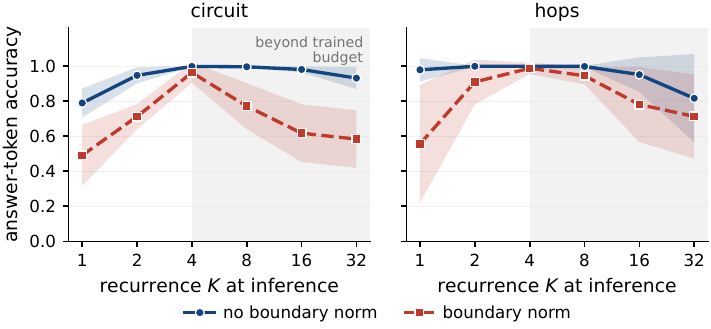}
\caption{Models trained from scratch with the loop topology fixed before training, on
\texttt{circuit} (left) and one cell of \texttt{hops} (right); means over \ScrSeedsA{}
seeds, bands one standard deviation. Both configurations share $r_0=4$, so the comparison
at equal $K$ is free of the overshoot confound of Sec.~\ref{sec:r4}. The curve with a
boundary normalisation peaks at its trained depth and falls away on both sides; the curve
without one stays above \ScrOffKTwo{} from $K=2$ to $K=16$.}
\label{fig:axisa}
\end{figure}

Two series of fifty training runs (32 on the normalisation axis, 18 on the depth axis;
\ScrSeedsA{} seeds per configuration here), one on \texttt{circuit} and one on a single
cell of \texttt{hops} ($d^\star=1$, width 8; App.~\ref{app:scope} says why one cell). All
runs are competent by the pre-declared gate, with answer-token accuracy between 0.99 and
1.00 at the end of training. The two configurations differ only in whether the state is
normalised at the end of every loop step and share the trained budget $r_0=4$. We
describe \texttt{circuit} first.

The two are closest at exactly one depth, the one they were trained at:
\ScrOffKFour{} without boundary normalisation against \ScrOnKFour{} with it,
$p=\ScrPKFour$ by the exact rank test over seeds (Table~\ref{tab:axisa}). The
normalisation therefore does not prevent learning. At every other depth the gap is wider,
and the arm with normalisation falls away in both directions: \ScrOnKTwo{} at half the
trained depth and \ScrOnKThirtyTwo{} at eight times it, against \ScrOffKTwo{} and
\ScrOffKThirtyTwo{} (Fig.~\ref{fig:axisa}). Below the budget both arms lose accuracy and
the normalised one loses more: at $K=1$, a single application of a six-layer block,
\ScrOnKOne{} against \ScrOffKOne{} ($p=\ScrPKOne$).

Beyond the trained budget the normalised arm falls monotonically, to \ScrOnKEight{},
\ScrOnKSixteen{} and \ScrOnKThirtyTwo{} at two, four and eight times the budget, while the
configuration without normalisation holds \ScrOffKEight{}, \ScrOffKSixteen{} and
\ScrOffKThirtyTwo{}; the gap is widest at four times the budget. The effect is an
interaction with the number of loop steps, and it separates the two sets of seeds almost
completely, with Cliff's $\delta$ reaching $\ScrDeltaMax$; each of the \ScrBeyondCells{}
depths beyond the budget is significant on its own ($p$ from $\ScrBeyondPMin$ to
$\ScrBeyondPMax$) and \ScrBeyondHolm{} of the \ScrBeyondCells{} survive the Holm
correction. The shape of the two curves is the result we rely on: the configuration with
boundary normalisation peaks at the depth it was trained at and degrades on both sides of
it, while the configuration without it holds above \ScrOffKSixteen{} from the trained
depth to four times above it.

The second task reproduces the shape with a smaller effect beyond the budget
(Fig.~\ref{fig:axisa}, right; Table~\ref{tab:axisatwo}). On \texttt{hops} the arms are
indistinguishable at the trained depth (\ScrTwoOffKFour{} against \ScrTwoOnKFour{},
$p=\ScrTwoPKFour$) and the normalised arm again falls away on both sides of it:
\ScrTwoOnKOne{} against \ScrTwoOffKOne{} at $K=1$ ($p=\ScrTwoPKOne$) and
\ScrTwoOnKEight{} against \ScrTwoOffKEight{} at twice the budget
($\delta=\ScrTwoDeltaMax$, $p=\ScrTwoPMin$). The gap is still significant at four times
the budget ($p=\ScrTwoPKSixteen$) and no longer at eight ($p=\ScrTwoPKThirtyTwo$);
\ScrTwoBeyondHolm{} of the \ScrTwoBeyondCells{} depths beyond the budget survive Holm.
The sign holds on both tasks and the size beyond the budget is task-dependent.

The direction of this effect is the main result of this half of the study. A naive
reading of Sec.~\ref{sec:r1} would suggest that boundary normalisation is useful,
because removing it breaks Ouro. Here, where the topology is fixed before training, the
models that never saw normalisation inside the loop hold their accuracy over a wider
range of $K$. Agreement explains Sec.~\ref{sec:r1}, since a model learns a readout for
the loop it has and what hurts it is a change to that loop, whatever the direction. It
does not by itself explain the present section: both configurations here run with the
loop they were trained with, and one still degrades beyond its budget. Agreement is
therefore necessary and not sufficient, and the normalisation separates two arms that
both satisfy it. The state norm points to how. The configuration without boundary normalisation lets its
state norm grow across loop steps by a factor of $4.84$, from $366$ to $1772$, and holds
its accuracy almost unchanged across $K$; the one with normalisation holds the norm nearly
constant, a factor of $1.45$, and degrades. If norm growth were the cause of failure, the
sign of this relation would be the other way round. A reading consistent with the sign is
that growth of the residual stream damps the loop: the block's output scale is set by its
internal normalisations, so as the state grows each application moves it by a smaller
relative amount and the iteration settles, whereas a boundary normalisation rescales the
state before every step, so the updates keep their full relative size and a readout
calibrated at $r_0$ sees a state that has not settled. We did not measure the per-step
update size and offer this as the interpretation the norms support, in line with the
fixed-point analyses cited above, not as a demonstrated mechanism.

\subsection{The damage sits in the last block layer and the head}
\label{sec:refit}

If the readout is what the operation breaks, refitting the readout against the modified
loop must recover accuracy without touching the computation. That test needs weights we
control, so we run it on the models of Sec.~\ref{sec:r3} trained with boundary
normalisation, on which the same operation is available. We freeze the loop, untie the
unembedding from the input embedding, and train only it and the final normalisation for
300 steps, with no gradient entering the loop. At the trained depth this recovers \RefitOneFrac{}\% of the accuracy
lost on \texttt{circuit} (\RefitOneSurg{} to \RefitOneRefit{}, against \RefitOneBase{} as
trained, $p=\RefitOnePSurg$ over \RefitOneSeeds{} seeds) and \RefitTwoFrac{}\% on
\texttt{hops} (\RefitTwoSurg{} to \RefitTwoRefit{}, $p=\RefitTwoPSurg$), and stays below
the model as trained on both ($p=\RefitOnePBase$, $p=\RefitTwoPBase$; exact signed-rank
over seeds). The recovered third is a plateau:
3000 steps move it from \RefitCurveHeadNormThreeHundred{}\% to
\RefitCurveHeadNormThreeThousand{}\% on \texttt{circuit}, and the unembedding alone gives
the same curve (Table~\ref{tab:refitcurve}). Unfreezing in addition the last layer of the
recurrent block, shared across loop steps and holding 4.2M of the 19M parameters, returns
\RefitCurveHeadNormLastThirty{}\% of the loss within 30 steps and
\RefitCurveHeadNormLastThreeThousand{}\% by 3000, and on \texttt{hops} 98\% within 30.
The procedure is not free: at $K=1$, where the operation is an identity, the retrained
head still gains \RefitOneRefitOne{} over \RefitOneSurgOne{}, so part of the third is
extra training and the plateau is the robust finding. The operation therefore
damages the block's last layer together with the head that reads it, and this pair has to
agree with the loop. Beyond the trained budget the single-point refit is task-dependent
(Table~\ref{tab:refit}): on \texttt{hops} the refitted readout matches the model as
trained, while on \texttt{circuit} the model as trained has itself fallen to
\RefitOneBaseSixteen{} by $K=16$.

\subsection{Looped against stacked depth at matched overshoot}
\label{sec:r4}

We fix the total number of layer applications per token at 24 and split it three ways:
four loop steps of six layers, two of twelve, and one of twenty-four, the last an
ordinary transformer without a loop. All three are trained without boundary
normalisation; the four-step loop is the no-normalisation arm of Sec.~\ref{sec:r3} on a
separate set of six seeds. Compared at equal $K$ the result is a tidy ordering in favour
of looping, with \BdeepFar{}, \BmidFar{} and \BwideFar{} at $K=32$ (Fig.~\ref{fig:axisb},
left). That ordering is an artefact. The one-step configuration has a trained budget of
$r_0=1$, so $K=32$ is a thirty-two-fold overshoot for it against an eight-fold overshoot
for the four-step configuration, and the left panel compares configurations that have
left their budgets by very different distances. Since all three apply 24 layers at their
trained budget, equal overshoot is also equal inference compute (layers~$\times K =
24\,K/r_0$): the fixed-$K$ comparison confounds overshoot with a four-fold difference in
compute, and matching removes both at once. At matched overshoot
the ordering is no longer detectable on \texttt{circuit} (Fig.~\ref{fig:axisb},
Table~\ref{tab:axisb}): the three configurations lie within one standard deviation across
seeds of each other, at \BOneDeepMultEight{}, \BOneMidMultEight{} and \BOneWideMultEight{}
at eight times the budget, and no pairwise test rejects. This is a statement about
detectability at \ScrSeedsB{} seeds: the exact Mann--Whitney test at that size rejects at
the 0.05 level only for $|\delta|\ge0.72$, and the observed $\delta=+0.33$ for the loop
over the stack at eight times the budget lies inside that range.

On \texttt{hops} the confound is the same but a gap survives it and shrinks without
closing (Table~\ref{tab:axisbtwo}): the $4\times6$ loop reaches \BTwoDeepMultEight{}
against \BTwoMidMultEight{} for $2\times12$ ($\delta=\BTwoMidDelta$, $p=\BTwoMidP$) and
\BTwoWideMultEight{} for the $1\times24$ stack ($\delta=\BTwoWideDelta$, $p=\BTwoWideP$).
We apply no multiplicity correction on this axis and read it as exploratory; across the
eight comparisons of Tables~\ref{tab:axisb} and~\ref{tab:axisbtwo} the one significant
difference would not survive Holm. Whether any advantage survives matched overshoot is
thus task-dependent.

\section{Discussion}

We read these results as a contract between the loop and its readout: extra loop steps
are usable only while the readout is calibrated for the loop it was trained with, and
they stay usable further beyond the budget when the loop settles as it iterates, which a
boundary normalisation prevents. The first clause is why ``remove the normalisation'' and
``train without it'' have opposite consequences; the second is why, of two configurations
that each agree with their own loop, the one without the normalisation extrapolates
further. The refit curve locates the contract in the last
block layer and the head that reads it. Two points follow for practice: extrapolation
beyond a trained budget is best reported against the overshoot $K/r_0$, since the
absolute axis confounds any comparison in which budgets differ; and a modification inside
the loop of a released model should be assumed to invalidate its readout until measured,
even when the state scale is intact.

\section{Limitations}
\label{sec:limits}

Scale and causality sit in different halves of the study: the causal half uses models of 19
to 76M parameters on two families of synthetic tasks, while the released models carry
scale but cannot have their training topology varied. The effect on the normalisation axis
is therefore established in sign in both halves and in magnitude only at the small scale.

Competence of the released models is limited. In the zero-shot pass only Ouro on
\texttt{circuit} clears the floor; four in-context examples raise \texttt{circuit} to
\OTSbaseCircuitPeak{} and \OFTbaseCircuitPeak{} and lift \texttt{hops} above both floors, to
\OTSbaseHopsPeak{} and \OFTbaseHopsPeak{} against a majority floor of
\OTSbaseHopsProbeMajority{}, but not to the pre-declared threshold of 0.30, which stands:
\texttt{hops} enters as a secondary observation that fails the gate, and the headline
result of Sec.~\ref{sec:r1} rests on \texttt{circuit} for both Ouro models; the collapse
statement is unaffected, since it concerns the structure of the prediction and not its
correctness. Huginn fails the gate on all three families (Table~\ref{tab:huginn}), so the
direction that adds a boundary normalisation to released weights remains untested. The
answer-token convention is fixed per tokeniser (Huginn merges the space and the digit
into one token, Ouro separates them); an earlier convention that scored the wrong token
for Huginn was caught by the in-alphabet rate, and the failure survives the correction,
so it is a competence limit under our prompt and not a scoring artefact. Accuracy is
scored by arg-max over the full vocabulary (Sec.~\ref{sec:metrics}); we do not report the
alphabet-restricted reading, recoverable from the logged top-5 tokens, and rely on the
format statistics of App.~\ref{app:collapse} to show that the released arms after the
operation have left the answer format. The causal half runs on two of the three
families; App.~\ref{app:scope} records why \texttt{paragg} and the multi-cell
\texttt{hops} grid are absent and that nothing transfers to deeper task instances.

The pre-registered primary cell was on the second evaluation tier and the few-shot pass
computed the first tier only, so the paired gap test is reported on \PairedTier{} at the
same $r=32$ with $n=\PairedOTScircuitN$ to $\PairedOTShopsN$ paired instances per pair;
direction and magnitude match the pre-registration, the number of cells and the sample
size do not. Evaluation instances come from the materialised grid where it holds the
trained cell and are otherwise generated with a disjoint index range (three of four
\texttt{circuit} cells, the \texttt{hops} cell). In the widest configuration without
boundary normalisation the state norm reaches 1772; our bfloat16 runs stayed stable, but
we claim only that norm growth does not explain the failure, not that removing the
normalisation is safe at any budget. We did not vary the readout normalisation, the axis
of \citet{sharma2026readout}.

\section{Conclusion}

Normalisation placement inside the loop is not a stability hyperparameter but a contract
between the loop and the readout. Removing it from released weights breaks the model
within a few loop steps while the state scale stays within a factor of \NormRatioMax{}:
the head leaves the answer alphabet and later commits to a constant that is not an answer.
A readout refit over a frozen loop recovers a third of the loss; retraining the last block
layer with it recovers all of it. Trained from scratch, the absence of the
normalisation is never worse than its presence and better away from the trained depth,
although both configurations agree with their own loop: agreement is necessary for depth
to be reusable, and given agreement the normalisation is a cost. At matched overshoot the
advantage of looping over a stack is not detectable at six seeds on one task and reduced
on the other.

\subsubsection*{Reproducibility statement}

The full code, with the protocol, task generators, verifiers, seeds and grids fixed in
a single module and hashed into every logged row, will be released publicly after the review process. All numbers in the text and in the tables are
generated from the run ledgers by one script, and no number is typed by hand. The
appendix lists the model configurations and training hyperparameters
(Tables~\ref{tab:models} and~\ref{tab:hyper}), the full accuracy grids
(Tables~\ref{tab:erelacc} and~\ref{tab:huginn}) and every per-depth table behind the
statistics in the text; the gate thresholds are in Sec.~\ref{sec:metrics} and the arms
that fail them are named in Sec.~\ref{sec:limits}.

\subsubsection*{AI use statement}

Generative AI tools were used for code review and for finding bugs, and for editing
this text. We did not use generative AI tools for any task with required disclosure:
not for generating data, developing the conceptual framework, formulating claims,
proposing or refining hypotheses, designing or giving feedback on the methodology or the
experiments, implementing the methods, cleaning or reformatting data, or interpreting
the results; translation and qualitative data analysis do not apply to this work.
Everything not listed above was done by the authors. All experiments were run by the
authors on their own hardware, and every number reported here is produced by the logged
runs rather than by a language model. We reviewed all AI-assisted work: every issue
raised in AI-assisted code review was checked and fixed by the authors, the protocol,
the answer-scoring convention, the floors, the pre-registered test and the gates were
checked by hand against the logs, and the identity check at $r=1$ served as a standing
correctness check on the implementation of the operation. We take responsibility for the
final content of this work, including the text, the claims and the artifacts.

\bibliography{paper}
\bibliographystyle{iclr2027_conference}

\appendix
\raggedbottom

\section{Appendix}

\subsection{Models}

\begin{table}[H]
\centering
\begin{tabular}{lllp{4.2cm}}
\toprule
role & model & parameters & loop topology \\
\midrule
released & Huginn-3.5B & 3.5B & no boundary norm measured \\
released & Ouro-1.4B & 1.4B & boundary norm, shared with readout \\
released & Ouro-2.6B & 2.6B & boundary norm, shared with readout \\
from scratch & LoopedLM, $d{=}512$ & 19--76M & either, dedicated, set before training \\
\bottomrule
\end{tabular}
\caption{Models. The released checkpoints carry the scale and the models trained from
scratch carry the causal claims; see Sec.~\ref{sec:models} for why the two halves answer
different questions.}
\label{tab:models}
\end{table}

\subsection{Training hyperparameters}

\begin{table}[H]
\centering
\small
\begin{tabular}{ll}
\toprule
setting & value \\
\midrule
width $d$, heads, MLP multiplier & 512, 8, 4 \\
layers $\times$ trained loop steps & $6\times4$ (normalisation axis and loop arm); $12\times2$; $24\times1$ \\
boundary normalisation & RMSNorm between loop steps, skipped on the last; on or off \\
readout normalisation & RMSNorm before the unembedding, always on \\
context, tokens per batch, training tokens & 256, 65\,536, $3\times10^{8}$ \\
optimiser, peak learning rate, precision & AdamW, $3\times10^{-4}$, bfloat16 \\
evaluation loop counts $K$ & 1, 2, 4, 8, 16, 32 \\
seeds & \ScrSeedsA{} on the normalisation axis, \ScrSeedsB{} on the depth axis \\
\bottomrule
\end{tabular}
\caption{Training hyperparameters of the models trained from scratch. All configurations
share data, tokeniser, schedule and seeds; they differ only in the row that names them.}
\label{tab:hyper}
\end{table}

\subsection{Huginn in the zero-shot pass}

\begin{table}[H]
\centering
\footnotesize
\setlength{\tabcolsep}{8pt}
\begin{adjustbox}{max width=\textwidth}
\begin{tabular}{llrrrrrrrrr}
\toprule
Huginn-3.5B & task & $r{=}1$ & $r{=}2$ & $r{=}4$ & $r{=}8$ & $r{=}16$ & $r{=}32$ & $r{=}64$ & $r{=}128$ & $r{=}256$ \\
\midrule
as released & circuit & 0.004 & 0.000 & 0.000 & 0.311 & 0.414 & 0.448 & 0.445 & 0.453 & 0.451 \\
as released & hops & 0.000 & 0.002 & 0.000 & 0.000 & 0.001 & 0.001 & 0.000 & 0.000 & 0.000 \\
as released & paragg & 0.100 & 0.000 & 0.035 & 0.171 & 0.018 & 0.044 & 0.059 & 0.060 & 0.055 \\
boundary norm added & circuit & 0.004 & 0.000 & 0.000 & 0.299 & 0.416 & 0.446 & 0.452 & 0.446 & 0.450 \\
boundary norm added & hops & 0.000 & 0.001 & 0.000 & 0.000 & 0.000 & 0.000 & 0.000 & 0.000 & 0.000 \\
boundary norm added & paragg & 0.100 & 0.000 & 0.039 & 0.203 & 0.023 & 0.050 & 0.054 & 0.051 & 0.053 \\
\bottomrule
\end{tabular}

\end{adjustbox}
\caption{The reverse operation on released weights: Huginn-3.5B as released and with a
boundary normalisation added, zero-shot, accuracy against the inference-time loop count on
all three families. Neither arm clears the competence gate of Sec.~\ref{sec:metrics} on any
family, which is why this direction carries no claim in the text.}
\label{tab:huginn}
\end{table}

\subsection{Scope of the causal half}
\label{app:scope}

The causal half runs on two of our three families, and we measured why the third is
absent. A single cell of \texttt{hops} is learned by an abrupt transition at around 2750
steps, about 700 thousand examples, matching the reported shape for induction heads
\citep{olsson2022}; an eight-cell mixture trained for the same number of steps, 128
thousand examples per cell, shows no transition in any cell. Since the mixture saw fewer
examples per cell than the single cell needed, these runs do not separate a per-cell
threshold from a cumulative one; they establish that the grid is not learned as a mixture
within our budget, which is why the causal half trains on one cell. A control task on the
same data path is learned within 250 steps, clearing the generator, the encoder, the answer
position and the loss; \texttt{paragg} is not solved by any configuration.

The same models bound the claim from the other side. On the \TransferCells{}
\texttt{hops} cells at $d^\star=2,3,4$ that no run saw in training, every configuration
scores between \TransferMin{} and \TransferMax{} against a chance floor of
\TransferFloor{}, averaged over inference depths, and raising $K$ does not help. Below the
floor is the signature of Sec.~\ref{sec:r2} again: on an instance it has not learned the
model commits to a wrong answer instead of spreading its mass. Loop depth at inference and
task depth are separate axes, and our results show no transfer from the first to the
second.

The control task is ``copy the starting letter''; it travels the same data path, format
and answer position as \texttt{hops} and is learned to accuracy one. A step is 256
sequences, and the transitioning cell is $d^\star=1$, width 8. \texttt{paragg} remains at
its floor in every configuration, including after a fix to its generator whose earlier
version produced a constant label at small widths; that earlier version is the reason we
verify every generator with independent code.

\subsection{Normalisation axis per depth on \texttt{circuit}}

\begin{table}[H]
\centering
\footnotesize
\setlength{\tabcolsep}{8pt}
\begin{adjustbox}{max width=\textwidth}
\begin{tabular}{lrrrrrr}
\toprule
configuration & $K{=}1$ & $K{=}2$ & $K{=}4$ & $K{=}8$ & $K{=}16$ & $K{=}32$ \\
\midrule
no boundary norm & 0.790 & 0.948 & 0.999 & 0.997 & 0.981 & 0.932 \\
boundary norm & 0.492 & 0.713 & 0.963 & 0.772 & 0.618 & 0.584 \\
\midrule
Cliff's $\delta$ & +0.94 & +1.00 & +0.52 & +0.84 & +0.87 & +0.86 \\
$p$ (Mann--Whitney) & $2.2\times10^{-7}$ & $3.3\times10^{-9}$ & $0.012$ & $1.1\times10^{-5}$ & $4.0\times10^{-6}$ & $5.3\times10^{-6}$ \\
\bottomrule
\end{tabular}

\end{adjustbox}
\caption{Models trained from scratch on \texttt{circuit}: answer-token accuracy
against the inference-time number of loop steps $K$, averaged over \ScrSeedsA{}
training seeds per configuration. Both configurations have six layers, four loop steps
during training and 19M parameters. Cliff's $\delta$ and the exact Mann-Whitney $p$ are
computed over seeds, with the seed as the unit of analysis.}
\label{tab:axisa}
\end{table}

\subsection{Normalisation axis per depth on \texttt{hops}}

\begin{table}[H]
\centering
\footnotesize
\setlength{\tabcolsep}{8pt}
\begin{adjustbox}{max width=\textwidth}
\begin{tabular}{lrrrrrr}
\toprule
configuration & $K{=}1$ & $K{=}2$ & $K{=}4$ & $K{=}8$ & $K{=}16$ & $K{=}32$ \\
\midrule
no boundary norm & 0.979 & 0.999 & 0.999 & 0.999 & 0.952 & 0.817 \\
boundary norm & 0.556 & 0.910 & 0.988 & 0.946 & 0.781 & 0.713 \\
\midrule
Cliff's $\delta$ & +0.88 & +0.53 & +0.20 & +0.84 & +0.67 & +0.37 \\
$p$ (Mann--Whitney) & $2.3\times10^{-6}$ & $0.010$ & $0.361$ & $8.9\times10^{-6}$ & $9.0\times10^{-4}$ & $0.080$ \\
\bottomrule
\end{tabular}

\end{adjustbox}
\caption{Models trained from scratch on one cell of \texttt{hops} ($d^\star=1$, width 8):
answer-token accuracy against the inference-time number of loop steps $K$, averaged over
\ScrTwoSeedsA{} training seeds per configuration. Evaluation instances for this cell are
generated held out with a disjoint index range, since the materialised evaluation grid
does not contain it.}
\label{tab:axisatwo}
\end{table}

\subsection{Collapse metrics}
\label{app:collapse}

At $r=32$ on \texttt{hops} the unmodified arms spread their answers over 25 and 30
distinct tokens, with 4.21 and 4.69 bits of mutual information between the input and
the arg-max prediction and a modal share of 0.20 and 0.11. With the boundary
normalisation removed, Ouro-1.4B returns a single token for the whole split at zero
mutual information and a modal share of 1.000, and Ouro-2.6B returns two tokens with a
modal share of 0.860. On the binary family, where a two-token prediction is not by
itself a collapse, the modal share rises from 0.50 and 0.54 in the unmodified arms to
0.85 and 0.83 after the rearrangement.

\subsection{Top-1 minus top-2 logit}

\begin{figure}[H]
\centering
\includegraphics[width=0.85\textwidth]{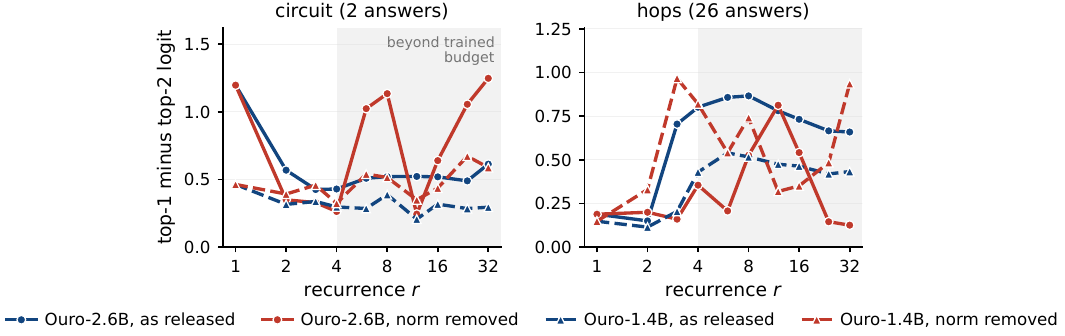}
\caption{Top-1 minus top-2 logit against the number of loop steps for both task
families. This quantity is noisy across $r$ and is interpretable only together with the
modal share; Sec.~\ref{sec:r2} cites it only at the trained budget and at $r=32$.}
\label{fig:margin}
\end{figure}

\subsection{Looped against stacked depth at matched overshoot}

\begin{figure}[H]
\centering
\includegraphics[width=0.85\textwidth]{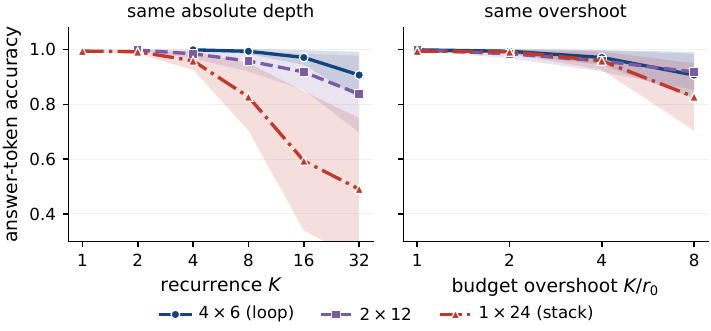}
\caption{Three placements of a matched depth of 24 layer applications per token, trained
from scratch on \texttt{circuit}. Lines are means over \ScrSeedsB{} seeds, bands one
standard deviation. Left: accuracy against the absolute loop count $K$, which is how
such comparisons are usually reported, with each configuration shown from its own trained
budget upwards. Right: the same runs against the budget overshoot $K/r_0$, restricted to
the overshoots all three configurations reach. The ordering on the left is absent on the
right; that difference is the point of Sec.~\ref{sec:r4}.}
\label{fig:axisb}
\end{figure}

\begin{table}[H]
\centering
\small
\begin{tabular}{lrrrr}
\toprule
configuration & $K/r_0{=}1$ & $K/r_0{=}2$ & $K/r_0{=}4$ & $K/r_0{=}8$ \\
\midrule
$4\times6$ (loop) & 0.999 & 0.994 & 0.971 & 0.907 \\
$2\times12$ & 0.998 & 0.985 & 0.958 & 0.918 \\
$1\times24$ (stack) & 0.995 & 0.992 & 0.959 & 0.828 \\
\midrule
$\delta$, loop vs $2\times12$ & +0.22 & +0.39 & +0.11 & -0.11 \\
$p$, loop vs $2\times12$ & 0.589 & 0.310 & 0.818 & 0.818 \\
\midrule
$\delta$, loop vs $1\times24$ (stack) & +0.50 & +0.25 & +0.25 & +0.33 \\
$p$, loop vs $1\times24$ (stack) & 0.180 & 0.589 & 0.589 & 0.394 \\
\bottomrule
\end{tabular}

\caption{Answer-token accuracy at matched budget overshoot $K/r_0$ for three
placements of 24 layer applications per token, averaged over \ScrSeedsB{} training seeds
on \texttt{circuit}. Each column compares the three configurations after leaving their
trained budgets by the same factor.}
\label{tab:axisb}
\end{table}

\begin{table}[H]
\centering
\small
\begin{tabular}{lrrrr}
\toprule
configuration & $K/r_0{=}1$ & $K/r_0{=}2$ & $K/r_0{=}4$ & $K/r_0{=}8$ \\
\midrule
$4\times6$ (loop) & 1.000 & 0.998 & 0.998 & 0.957 \\
$2\times12$ & 1.000 & 1.000 & 0.948 & 0.733 \\
$1\times24$ (stack) & 1.000 & 0.997 & 0.925 & 0.737 \\
\midrule
$\delta$, loop vs $2\times12$ & +0.00 & -0.17 & +0.58 & +0.53 \\
$p$, loop vs $2\times12$ & 1.000 & 0.699 & 0.132 & 0.180 \\
\midrule
$\delta$, loop vs $1\times24$ (stack) & +0.00 & +0.03 & +0.56 & +0.78 \\
$p$, loop vs $1\times24$ (stack) & 1.000 & 1.000 & 0.132 & 0.026 \\
\bottomrule
\end{tabular}

\caption{Looped against stacked depth on one cell of \texttt{hops}: answer-token accuracy at matched budget
overshoot $K/r_0$ for the three placements of 24 layer applications per token, averaged
over \ScrTwoSeeds{} seeds, with Cliff's $\delta$ and the exact Mann--Whitney $p$ against
the four-step loop. All three are trained without boundary normalisation; the four-step
loop is the no-normalisation architecture of Table~\ref{tab:axisatwo} on a separate set of
six seeds, and its higher value at eight times the budget lies within the seed spread of
Fig.~\ref{fig:axisa}. Unlike on \texttt{circuit} (Table~\ref{tab:axisb}), the loop retains
an advantage over the stack at eight times the budget ($p=\BTwoWideP$ over
\ScrTwoSeeds{} seeds, uncorrected), and none over $2\times12$ ($p=\BTwoMidP$).}
\label{tab:axisbtwo}
\end{table}

\subsection{Paired gap test in full}
\label{app:paired}

The two \texttt{circuit} pairs pass the competence gate and the two \texttt{hops} pairs
do not and are reported as observations. On \texttt{circuit} the gap moves from
${\PairedOTScircuitGapBase}$ to ${\PairedOTScircuitGapSurg}$ for Ouro-2.6B
($\Delta=\PairedOTScircuitDelta$, 95\% cluster bootstrap interval
${\PairedOTScircuitCI}$, McNemar $p=\PairedOTScircuitP$, $n=\PairedOTScircuitN$) and from
${\PairedOFTcircuitGapBase}$ to ${\PairedOFTcircuitGapSurg}$ for Ouro-1.4B; on
\texttt{hops} $\Delta$ is $\PairedOTShopsDelta$ for both models, to three decimals.

\begin{table}[H]
\centering
\footnotesize
\setlength{\tabcolsep}{3.5pt}
\begin{adjustbox}{max width=\textwidth}
\begin{tabular}{llrrrrr}
\toprule
model & task & $n$ & gap, as released & gap, surgery & $\Delta$ [95\% CI] & $p$ \\
\midrule
Ouro-1.4B & circuit & 800 & $-0.011$ & $+0.516$ & $-0.527$ $[-0.569, -0.490]$ & $1.5\times10^{-126}$$^{*}$ \\
Ouro-1.4B & hops & 900 & $-0.104$ & $+0.006$ & $-0.110$ $[-0.185, -0.055]$ & $4.9\times10^{-32}$$^{*}$ \\
Ouro-2.6B & circuit & 800 & $-0.101$ & $+0.490$ & $-0.591$ $[-0.609, -0.567]$ & $1.0\times10^{-140}$$^{*}$ \\
Ouro-2.6B & hops & 900 & $-0.049$ & $+0.061$ & $-0.110$ $[-0.222, -0.031]$ & $5.0\times10^{-29}$$^{*}$ \\
\bottomrule
\end{tabular}

\end{adjustbox}
\caption{The pre-registered paired test in full. The gap is the frozen probe minus the
model output on the same instance; the prediction fixed before the first forward pass was a
small gap for Ouro as released and a large one after the boundary normalisation is removed.
$\Delta$ is the difference of gaps with a cluster bootstrap interval over (family,
$d^\star$, data seed), $p$ is the exact McNemar test on paired instances, and an asterisk
marks rejection after the Holm correction.}
\label{tab:paired}
\end{table}

\subsection{Accuracy on the full grid}

\begin{table}[H]
\centering
\footnotesize
\setlength{\tabcolsep}{4pt}
\begin{tabular}{llrrrrrrrrrr}
\toprule
arm & task & $r{=}1$ & $r{=}2$ & $r{=}3$ & $r{=}4$ & $r{=}6$ & $r{=}8$ & $r{=}12$ & $r{=}16$ & $r{=}24$ & $r{=}32$ \\
\midrule
O26-base & circuit & 0.530 & 0.584 & 0.627 & 0.642 & 0.620 & 0.598 & 0.588 & 0.590 & 0.591 & 0.583 \\
O26-base & hops & 0.074 & 0.107 & 0.164 & 0.166 & 0.154 & 0.142 & 0.147 & 0.138 & 0.122 & 0.106 \\
O26-fbraw & circuit & 0.530 & 0.500 & 0.058 & 0.000 & 0.000 & 0.000 & 0.000 & 0.000 & 0.000 & 0.000 \\
O26-fbraw & hops & 0.074 & 0.071 & 0.049 & 0.082 & 0.038 & 0.000 & 0.000 & 0.000 & 0.000 & 0.000 \\
O14-base & circuit & 0.471 & 0.591 & 0.600 & 0.619 & 0.561 & 0.550 & 0.537 & 0.530 & 0.527 & 0.524 \\
O14-base & hops & 0.074 & 0.070 & 0.121 & 0.154 & 0.140 & 0.152 & 0.149 & 0.149 & 0.128 & 0.117 \\
O14-fbraw & circuit & 0.471 & 0.090 & 0.000 & 0.000 & 0.000 & 0.000 & 0.000 & 0.000 & 0.000 & 0.000 \\
O14-fbraw & hops & 0.074 & 0.027 & 0.000 & 0.000 & 0.000 & 0.000 & 0.004 & 0.007 & 0.000 & 0.000 \\
\bottomrule
\end{tabular}

\caption{Accuracy against the number of loop steps for the released arms with four
in-context examples, averaged over strata of $d^\star$ and $w$ with $n=100$ instances
per stratum. Arms named \texttt{-fbraw} have the boundary normalisation removed. The
trained budget is $r_0=4$.}
\label{tab:erelacc}
\end{table}

\subsection{Readout refit after the operation}
\label{app:refit}

\begin{table}[H]
\centering
\small
\begin{tabular}{lrrrrrr}
\toprule
circuit & $K{=}1$ & $K{=}2$ & $K{=}4$ & $K{=}8^{*}$ & $K{=}16^{*}$ & $K{=}32^{*}$ \\
\midrule
as trained & 0.578 & 0.742 & 0.998 & 0.864 & 0.621 & 0.601 \\
loop-boundary norm removed & 0.578 & 0.626 & 0.639 & 0.701 & 0.708 & 0.710 \\
+ readout refit & 0.671 & 0.718 & 0.756 & 0.765 & 0.728 & 0.729 \\
\midrule
recovered & --- & 78\% & 33\% & 39\% & --- & --- \\
$p$ vs removed & 0.031 & 0.062 & 0.031 & 0.031 & 0.625 & 0.625 \\
$p$ vs as trained & 0.031 & 0.312 & 0.031 & 0.094 & 0.562 & 0.562 \\
\bottomrule
\end{tabular}

\begin{tabular}{lrrrrrr}
\toprule
hops & $K{=}1$ & $K{=}2$ & $K{=}4$ & $K{=}8^{*}$ & $K{=}16^{*}$ & $K{=}32^{*}$ \\
\midrule
as trained & 0.385 & 0.875 & 1.000 & 0.980 & 0.880 & 0.810 \\
loop-boundary norm removed & 0.385 & 0.600 & 0.775 & 0.805 & 0.805 & 0.830 \\
+ readout refit & 0.415 & 0.825 & 0.845 & 0.860 & 0.870 & 0.870 \\
\midrule
recovered & --- & 82\% & 31\% & 31\% & 87\% & --- \\
$p$ vs removed & 0.906 & 0.062 & 0.062 & 0.125 & 0.062 & 0.062 \\
$p$ vs as trained & 0.906 & 0.031 & 0.062 & 0.094 & 1.000 & 0.844 \\
\bottomrule
\end{tabular}

\caption{Three arms on the same loop weights of the models trained from scratch with
boundary normalisation: as trained, after the normalisation is removed, and after
refitting the readout against the modified loop. The refit trains only the final
normalisation and the unembedding, untied from the input embedding, for 300 steps with the
loop frozen. Entries are medians over the first \RefitOneSeeds{} training seeds (11--66),
a subset of the seeds of Table~\ref{tab:axisa}, so the ``as trained'' row is a six-seed
median and differs from the sixteen-seed means there. ``Recovered'' is
the share of the accuracy lost to the operation that the refit returns, reported where the
operation costs accuracy at all. Both $p$ values are the exact signed-rank test over
seeds, paired by seed: against the arm with the normalisation removed, and against the arm
as trained. An asterisk marks depths beyond the trained budget $r_0=4$.}
\label{tab:refit}
\end{table}

\subsection{Dose: boundary normalisation every $k$ loop steps}
\label{app:dose}

\begin{table}[H]
\centering
\footnotesize
\setlength{\tabcolsep}{8pt}
\begin{adjustbox}{max width=\textwidth}
\begin{tabular}{lrrr}
\toprule
boundary norm applied & $r{=}4$ & $r{=}8^{*}$ & $r{=}32^{*}$ \\
\midrule
every step (as released) & 0.642 & 0.598 & 0.583 \\
every 2nd & 0.496 & 0.489 & 0.481 \\
every 4th & 0.000 & 0.000 & 0.198 \\
every 8th & 0.000 & 0.000 & 0.000 \\
last step only & 0.000 & 0.000 & 0.000 \\
\midrule
entropy, every step (as released) & 0.95 & 0.93 & 1.49 \\
entropy, every 2nd & 3.51 & 2.56 & 7.19 \\
entropy, every 4th & 12.11 & 11.47 & 13.02 \\
entropy, every 8th & 12.11 & 12.00 & 14.17 \\
entropy, last step only & 12.11 & 12.00 & 5.67 \\
\midrule
$p$ vs as released, every 2nd & $3.1\times10^{-9}$ & $1.2\times10^{-5}$ & $1.8\times10^{-4}$ \\
$p$ vs as released, every 4th & $1.5\times10^{-154}$ & $1.0\times10^{-143}$ & $1.2\times10^{-45}$ \\
$p$ vs as released, every 8th & $1.5\times10^{-154}$ & $1.0\times10^{-143}$ & $4.2\times10^{-140}$ \\
$p$ vs as released, last step only & $1.5\times10^{-154}$ & $1.0\times10^{-143}$ & $4.2\times10^{-140}$ \\
non-increasing along the dose & yes & yes & yes \\
\bottomrule
\end{tabular}

\end{adjustbox}
\caption{Ouro-2.6B on \texttt{circuit} with four in-context examples, the boundary
normalisation applied every $k$-th loop step and always on the last, so that the readout
path is identical across rows. Rows are ordered by distance from the trained topology.
Accuracy is over $n=800$ instances per column, eight strata of 100; $p$ is the exact
McNemar test against the
model as released on the same instances, Holm-corrected within each $r$. Where $k\ge r$
the schedule reduces to normalising the last step only and the rows coincide exactly,
and this is checked. An asterisk marks depths beyond the trained budget.}
\label{tab:dose}
\end{table}

\subsection{Refit curve: steps and scope}
\label{app:refitcurve}

\begin{table}[H]
\centering
\small
\begin{tabular}{lrrrrrr}
\toprule
circuit, refit steps & 0 & 30 & 100 & 300 & 1000 & 3000 \\
\midrule
unembedding, accuracy & 0.639 & 0.729 & 0.744 & 0.751 & 0.755 & 0.761 \\
unembedding, recovered & 0\% & 25\% & 29\% & 32\% & 33\% & 34\% \\
+ final norm, accuracy & 0.639 & 0.726 & 0.742 & 0.756 & 0.756 & 0.765 \\
+ final norm, recovered & 0\% & 25\% & 29\% & 33\% & 33\% & 35\% \\
+ last loop block, accuracy & 0.639 & 0.791 & 0.869 & 0.940 & 0.982 & 0.996 \\
+ last loop block, recovered & 0\% & 43\% & 65\% & 85\% & 96\% & 100\% \\
\bottomrule
\end{tabular}

\vspace{4pt}
\begin{tabular}{lrrrrrr}
\toprule
hops, refit steps & 0 & 30 & 100 & 300 & 1000 & 3000 \\
\midrule
unembedding, accuracy & 0.775 & 0.845 & 0.855 & 0.845 & 0.870 & 0.865 \\
unembedding, recovered & 0\% & 31\% & 36\% & 31\% & 42\% & 40\% \\
+ final norm, accuracy & 0.775 & 0.845 & 0.860 & 0.845 & 0.870 & 0.860 \\
+ final norm, recovered & 0\% & 31\% & 38\% & 31\% & 42\% & 38\% \\
+ last loop block, accuracy & 0.775 & 0.995 & 1.000 & 1.000 & 1.000 & 1.000 \\
+ last loop block, recovered & 0\% & 98\% & 100\% & 100\% & 100\% & 100\% \\
\bottomrule
\end{tabular}

\caption{Recovery of the accuracy lost to the operation, by number of refit steps and by
what is unfrozen: the unembedding alone; plus the final normalisation; plus the last loop
block, i.e.\ the last layer of the recurrent block, shared across loop steps (4.2M
parameters). Loop weights below it stay frozen throughout. Medians over the first
\RefitOneSeeds{} seeds at the trained depth; ``recovered'' is the share of the loss
returned, computed from unrounded medians. Step 0 is the untouched head and coincides
with the arm after the operation by construction, checked per seed. The readout alone plateaus at a third on both
tasks; with the last block the loss is recovered in full.}
\label{tab:refitcurve}
\end{table}

\end{document}